# Opacity as Authority:
# Arbitrariness and the Preclusion of Contestation

**Naomi OMEONGA wa KAYEMBE**
Researcher in Cognitive Psychology and AI Ethics
Member, British Psychological Society (BPS)
Université de Nantes | naomi.omeongawakayembe@univ-nantes.fr

## Abstract

This article redefines arbitrariness not as a normative flaw or a symptom of domination, but as a foundational functional mechanism structuring human systems and interactions.
Diverging from critical traditions that conflate arbitrariness with injustice, it posits arbitrariness as a semiotic trait: a property enabling systems—linguistic, legal, or social—to operate effectively while withholding their internal rationale. Building on Ferdinand de Saussure's concept of l'arbitraire du signe, the analysis extends this principle beyond language to demonstrate its cross-domain applicability, particularly in law and social dynamics.

The paper introduces the "Motivation → Constatability → Contestability" chain, arguing that motivation functions as a crucial interface rendering an act's logic vulnerable to intersubjective contestation. When this chain is broken through mechanisms like "immotivization" or "Conflict Lateralization" (exemplified by *"the blur of the wolf drowned in the fish"*), acts produce binding effects without exposing their rationale, thus precluding justiciability.
This structural opacity, while appearing illogical, is a deliberate design protecting authority from accountability.

Drawing on Shannon's entropy model, the paper formalizes arbitrariness as $A = H(L|M)$ (conditional entropy). It thereby proposes a modern theory of arbitrariness as a neutral operator central to control as well as care, an overlooked dimension of interpersonal relations. While primarily developed through human social systems, this framework also illuminates a new pathway for analyzing explainability in advanced artificial intelligence systems.

# Opacity as Authority:
# Arbitrariness and the Preclusion of Contestation

Across the critical disciplines that have most incisively explored the architecture of social life, arbitrariness has emerged as a key signpost of symbolic domination and normative instability.

In political psychology, Social Dominance Theory (Sidanius & Pratto, 1999) offered a foundational account of how societies maintain group-based hierarchies through culturally contingent distinctions. Their notion of "arbitrary-set" bases of hierarchies, such as those organized by race, religion, or class, provided a powerful framework to understand how privilege becomes institutionalized through ideological legitimation instead of pure necessity.

In sociology, Pierre Bourdieu's theory of *Violence Symbolique* redefined the terrain of domination by shifting attention from coercive power to invisible inculcation (Bourdieu, 1977). Arbitrariness, in his formulation, is less imposed from above than lived as a necessity from within. Through processes of *Habitus* development and the misrepresentation of power-infused distinctions as natural, social agents come to reproduce structures that they themselves did not plan. Bourdieu does not simply critique inequality; he reveals how arbitrariness, once embedded, becomes the medium of consent, invisibly orchestrating both individual compliance and social reproduction. It is this generative capacity to order without justification that renders symbolic violence so potent, and so difficult to see.

Extending this logic beyond individual dispositions, arbitrariness also manifests in the broader inertia of institutional norms. These norms, once enacted, acquire a form of performative persistence. Not because they are justified, but because they are reiterated across contexts that preclude their contestation. Instead of the origin of the norm, what matters is its uptake and unexamined stability. In this sense, the misperception of what is arbitrary for what is natural, happens to be the reason why arbitrariness operates beneath the threshold of accountability. Throughout Michel Foucault's thinking, arbitrariness takes on its most expansive, genealogical force. Notably, in *Discipline and Punish* (Foucault, 1977) and his lectures at the Collège de France, Foucault traces the historic sedimentation of arbitrariness within institutions that present themselves as rational, such as law and medicine. His analyses reveal that norms do not emerge from truth, but from the tactical interplay of power and knowledge. What we take as necessary (e.g., the rules of legal procedure, the categories of psychiatric diagnosis, the architecture of the prison) are in fact the aftereffects of contingent struggles and institutional investments. For Foucault, what defines arbitrariness is the concealed engine of modern rationalities.

If arbitrariness has served as a symptom of domination or a signal of ideological infection in French sociology and philosophy, it is in the pioneering work of Claude Lévi-Strauss that we find its first formal articulation as a structural condition.

Lévi-Strauss (1963), in his analysis of myth and classification, understood that systems require oppositional pairings whose associations are not justified, but operationally efficient. The raw and the cooked, the male and the female, the sacred and the profane: these pairings work as they allow symbolic structure to unfold, not because they are true. Lévi-Strauss' findings support that arbitrariness is not a glitch but a generative device, necessary for the transmission and regulation of culture. He showed that the very stability of meaning depends on associations that are not derivable from nature and function, precisely, insofar as they are collectively asserted and assertive.

While these traditions diverge in method, object, and disciplinary lineage, they converge around a shared preoccupation: the enduring tension between hierarchy/leadership and performance/competence, between norm and legitimacy, between social form and moral consequence. Whether approached through social reproduction or systemic self-reference, arbitrariness consistently emerges as the sign of something deeper, a residue of power enacted without explanation, of structure stabilized without derivation, of authority asserted without justification. It is invoked when merit appears disconnected from status, when decision detaches from deliberation, or when institutions operate with impunity. And yet, across this shared terrain, arbitrariness remains curiously elusive: more often intuited than defined, more easily named than identified, and distributed across critiques rather than unified as an object of inquiry.

Each theoretical strand has offered its own vocabulary to make sense of the scenes where domination obscures its logic, or normativity reproduces without checking itself. But what binds these insights besides their brilliance, is their common reliance on a concept that has yet to be fully theorized. Arbitrariness appears in these social science accounts as both causes and cues, both tool and trace, but hardly as a structure capable of being tracked across domains.
This article proposes that arbitrariness may in fact constitute a consistent pattern: one that operates in systems, language, law, and social interaction alike, and that shapes not only how systems reproduce, but how acts become perceivable, opposable, or endured.

What follows is an attempt to bring arbitrariness into conceptual focus, not by contesting its prior uses, but by offering a framework through which their shared logic might finally become visible.

Such cross-domain transfers remain rare in social and human sciences, where epistemological caution tends to discourage transposition. But in fields such as physics, information theory, or systems science, the abstraction of concepts like singularity, symmetry, or feedback has long been recognized as methodologically valid. Indeed, these notions capture something essentially invariant beneath form or material differences.
In that same spirit, I treat arbitrariness not as a sociological category, nor a philosophical trope, but as a semiotic trait, a property that can help us track how acts enforce and bind, while remaining unreachable.

To begin modeling arbitrariness, I did not return to the political sciences of Machiavelli or Aristotle. Instead, I immersed myself in the classics of humanities, especially the Swiss author who pioneered modern linguistics: Ferdinand de Saussure and his notion of *l'arbitraire du signe* ("the arbitrariness of the sign").

When he first introduced his linguistic concept, Ferdinand de Saussure described the non-motivated relationship between signifier and signified as the foundational condition of symbolic language. The sign, as the vehicle of both signifier and signified, functions not because it is grounded in natural or logical necessity, but because it is socially enacted and defined by systemic contrast. As Saussure explains, "the bond between the signifier and the signified is arbitrary," emphasizing that language depends on differentiation and collective recognition instead of intrinsic logic within the symbol (Saussure, 1916/2011).

This mechanism of meaning formation extends beyond language and, from this localized observation, we may retain a rule: wherever a system generates stable, recognizable effects without inferential justification, we are in the domain of arbitrariness. Arbitrariness, in this strong sense, names the non-derivable that nevertheless structures. It does not need to make sense in order to produce sense. It enables shared worlds without referring to a foundational truth, but by collective enforcement, display, and systemic identification.

The transfer of regularities extracted from a domain-specific functional mechanism to explore cross-domain applicability is neither trivial nor metaphorical. The re-abstraction isolates the clauses and conditions of Saussurean arbitrariness, i.e., operational efficacy without internal justification, and tests it wherever systems produce reliable outcomes while withholding their internal rationale. What Saussure formalized in linguistics therefore allows us to illuminate an epistemic condition and an emergence threshold: the point at which systems function precisely because they do not explain themselves.

By adopting this logic of abstraction, arbitrariness exhibits the same latent structural commonalities across various fields such as law, normativity, legitimacy, and institutional design. The arbitrary link between sound and sense, once confined to language, becomes a cross-disciplinary analytical lens. Arbitrariness does not apply to the sign alone, it governs how we interpret, how we comply, and how we are bound. The question, then, is not whether arbitrariness occurs, but whether we are ready to model its structure and predict its effects.

This paper argues that arbitrariness should not be treated merely as a sign of domination, but as a generalizable condition: a functional mechanism that enables systems (linguistic, IT, legal, or social) to operate effectively without exposing their internal rationale. By shifting the analytic unit from the sign to the enacted sense, I show how arbitrariness governs not only how meaning is produced, but how actions are rendered incontestable. In many critical traditions, arbitrariness is conflated with injustice and assumed to be a failure of fairness or evidence of discretionary abuse. This article, by contrast, reframes arbitrariness not as a normative flaw, but as a design feature: a point at which systems bind, affect, or regulate through a neutralized channel marked by a possibility of constatation (*"constatability"*) that is arrested, a motive that is nowhere to be found. It is within this sealed structure that arbitrariness produces its effects.

And those effects are not univocal: arbitrariness may shield unaccountable harm, but it can also carry unclaimed care. The prosocial or offensive orientation of the act says nothing compared to the inaccessibility of its logic, i.e., the fact that it remains operative while resisting confrontation, explanation, or redress. Recognizing this reframes societal debates around the stakes of motivational legibility and justiciability.

This study develops an interdisciplinary theory of arbitrariness. Though its principles are herein restored primarily in terms of law and cognition, in the future, they may ultimately inform more technical inquiries such as explainability in advanced artificial intelligence systems.

## 1. Arbitrariness in Social Institutions: Binding With Organic Necessity

### 1.1. Arbitrariness in Language: The Sign as a Stabilized Disjunction

In his general course of linguistics, Saussure (1916/2011) famously declared that the bond between the *signifiant* (signifier) and the *signifié* (signified) is arbitrary.
This arbitrariness does not imply chaos, nor does it suggest that the relationship is irrational. Although it is systematic and socially regulated, it is not derivational. The connection between the sound-image "tree" and the concept of "tree" has no intrinsic, natural, or causal foundation: it works because it is shared.

There is no natural, logical, or derivational reason why a particular sound-image (e.g., "tree") should correspond to a particular concept. The link holds only by convention, not by any inner justification. What stabilizes this system, then, is a relational distinction and not an internal logic. Each sign functions not by anchoring meaning in essence, but by differing from other signs. It is the contrastive structure, not the intrinsic content, that grants the sign its efficacy. Although the sign has a real effect, although it signifies and means something, its basis is structurally unjustifiable: it cannot be defended in causal or natural terms. It operates, and continues to do so, because locutors accept it as such.

This insight introduces a deeper concept: arbitrariness is not randomness. It is not disorder or unpredictability. On the contrary, it is order without derivation. It refers to a system's ability to function effectively while remaining opaque to foundational explanation. It is a structural dissociation between efficiency and explicability. As such, arbitrariness is a condition in which a system remains operationally stable whereas its internal justifications are inaccessible, unavailable, or non-existent. It is not irrational; it is sub-relational: it occupies a space where effect precedes explanation, where functioning is decoupled from causal transparency.

This understanding is what allows us to abstract the general principle I have already mentioned: "wherever an ensemble produces stable, recognizable effects, yet lacks derivational justification, we are in the domain of arbitrariness". And this principle, though classically recognized in language, has broader applications.

This non-motivated linkage is what enables language as a system. If each sign were determined by some internal necessity or natural resemblance to its referent, symbolic communication would collapse into indexicality or mimesis. The arbitrariness of signs allows for difference, opposition, and combinatory play: the very elements that make linguistic meaning dynamic, generative, and iterable. It is precisely because signs are arbitrary that a limited set of phonemes can produce an infinite array of utterances, meanings, and social effects.

Once again: the arbitrary is not the random. It is the *non-derivable* that nevertheless structures. It is that which does not need to make sense in order to have sense-making power. It generates reality through *shared enactment*, not through logical derivation.
The sign works because it has been collectively accepted, not because it reflects a deeper truth or necessity.

Reformulated otherwise, what is virtually referred to as the "natural necessity" of the link between signifier and signified is better empirically grasped as an "organic effect" of social systems themselves. What is misidentified as either a natural imperative or a dispensable social construct is, in fact, a socially stabilized form of necessity, organic in essence and arbitrary in origin. In this view, the absence of a *natural* foundation does not entail the absence of necessity, and rather shows that natural necessity may be seen as a social imperative, based on shared conventions sustained through daily enactment that enable the ongoing production and normal functioning of our human systems.

### 1.2. Arbitrariness in Human Systems: The Sense as Imperative Functionality

Claude Shannon (1949) formalized this idea in his theory of communication, which separates the question of meaning from the engineering problem of transmission. For Shannon, communication succeeds when the signal received matches the signal sent, regardless of whether the content is interpreted identically. Semantic congruence is irrelevant. What matters is output reliability, not internal justification. Arbitrariness, here, becomes a condition of systemic performance.

Roland Barthes (1964/1977), extending the semiotic model into cultural analysis, showed that entire regimes of meaning (food, fashion, national identity) operate through signs whose logic, instead of being inferred, is enacted through repetition and social recognition.
Barthes' myth does not deceive; it distorts. It appears natural to the group while remaining entirely conventional. The power of these systems lies in their ability to stabilize meaning without exposing the terms of their construction.

A similar motif appears in early cybernetics. In a foundational paper, Rosenblueth, Wiener, and Bigelow (1943) describe black-box systems as those with (i) internal architecture opacity, and (ii) a behavior that is both purposeful and regulated by feedback. These systems are understood through their outputs. Though its internal mechanisms may be unknown, the black-box system performs. What matters is functional reliability, not transparency.

This principle finds one of its most mature social articulations in the work of Niklas Luhmann (1984/1995). Where Lévi-Strauss (1963) uncovered symbolic pairing, Luhmann formalized the operation of meaning-production as self-referential differentiation. In his theory of autopoietic systems, Luhmann proposes that legal, political, and institutional forms reproduce themselves not through recourse to external justification, but through recursive communication.

A legal system, for instance, does not require a justified rationale for each norm; it achieves legitimacy through procedural consistency and system-internal iteration. The empirical grounding of every norm is unnecessary as long as the system can continue to process itself. Accordingly, arbitrariness is not just tolerated: it is essential to operational closure. Systems need not explain their codes; they only need to distinguish and re-enter them. This is how Luhmann carried forward Lévi-Strauss's precocious intuition into a theory of self-sustaining fluency that has reshaped how we think about complexity and normative *epistēmē*[1].

Taken together, these findings converge on a structural regularity that crosses domain boundaries. Systems can produce observable, stable, and normatively effective outputs without exposing their internal logic. Arbitrariness, in this light, is not a systemic shortcoming. It is a foundational condition of our symbolic (linguistic), institutional (political), and relational (interpersonal) life.

It is the means whereby meaning stabilizes in the absence of causal transparency.
And this sense-making mechanism is the norm; never an exception. It is the always silent rule behind the visible one.

### 1.3. Arbitrariness in Legal Proceedings: Binding With Legal Rationale

Where the rule of law applies, legal systems provide a particularly rich terrain for examining the structural dynamics of arbitrariness (Costa et al., 2021). In addition to their social and institutional effects, legal decisions are binding by definition. However, on top of their effectiveness, the mode by which legal acts become publicly legitimate counts.

This is where the requirement of *motivation* appears: a decision must be "motivated", that is, it must expose the logic upon which it rests. This is not an ethical ideal; it is a procedural necessity that anchors the possibility of justice or *justiciability*: the ability for a decision to be contested, appealed, revised (Bressman, 2007; Costa et al., 2021).

---

[1] *Epistēmē* (ἐπιστήμη), in its Ancient Greek origin, refers not to "knowledge," but to the internal structure of knowability, as the stabilized conditions under which truth, validity, and justification are rendered possible. This usage contrasts with *doxa* (δόξα), or opinion, and aligns more closely with structured systems of intelligibility. In this article, *normative epistēmē* designates the self-sustaining architectures of authority through which systems generate not only meaning, but prescriptive force enabling obligations, classifications, or exclusions to emerge without external justification. This usage finds partial echo in Michel Foucault's deployment of *épistémè* in *Les Mots et les Choses* (1966), where he describes the historically contingent conditions of possibility for knowledge formation. However, while Foucault traces epistemic regimes genealogically as dispersed power/knowledge constellations, my approach seeks to formalize the threshold logic by which such regimes become non-contestable, and to map their emergence across domains where epistemic opacity carries out structural arbitrariness.

As with the linguistic sign, an absent motivation does not inhibit the function: the legal act remains binding, operational and institutionally enforced. However, such a legal decision has now acquired the character of arbitrariness in the specific sense defined in this paper: producing predictable and recognizable effects without publicly 'constatable' logic (Krygier, 2011).

A legal decision is structurally arbitrary when it is not accompanied by a motivation that renders its internal logic observable, spontaneously objectifiable, or publicly contestable.
Some legal decisions are supposed to be discretionary: this is not to be confused with "arbitrary". The discretionary act is legal exactly for the reason why the arbitrary decision is not (Cohen, 2015): the former is legally authorized to bypass motivation, when the latter is not and evades the very requirements meant to constrain it (Hart, 1961; Fuller, 1969).
A legally arbitrary act seeks to produce enforceable consequences without exposing the motives that could lead to its own 'defeatability', thus insulating itself from rational scrutiny (Bressman, 2007; Krygier, 2011). The arbitrariness of a legal act is not linked to its contradiction with a superior binding rule, but to its sheer lack of legal rationale. This absence of *motivation* can in turn affect the correctness and the legal conformity of the administrative or judicial decision (McDonald, 2015).

In matters governed by the rule of law, arbitrariness emerges when the chain from motivation to contestability is broken. A legal decision is not justiciable by means of its content; a legal decision becomes justiciable by the logic it explicitly makes observable. Motivation is what renders a decision's internal rationale externally visible, it transforms a silent authority into an articulated reason (Ho, 2000). Once made visible, that logic becomes constatable: it can be identified, interpreted, and situated within a broader normative landscape (Cananea & Parona, 2024). And only then does it become contestable, open to appeal, revision, or contradiction (Wade & Forsyth, 2014).

The sequential conditions under which a legal act qualifies as non-arbitrary can be synthesized as follows:

**Motivation → Constatability → Contestability**

When this chain is severed, when a legal act is executed without motivation, or with such vague or opaque reasoning that its logic cannot be assessed, the act becomes structurally arbitrary (Fuller, 1969; Vermeule, 2020). Its effects remain intact meanwhile its rationale cannot be interrogated. It is the presence of effects without the exposed articulation of their cause that defines legal arbitrariness.

Procedural irregularity is not the criterion; the interruption of the motivational sequence is (Luhmann, 1984/2004; *Motor Vehicle Mfrs. Ass'n v. State Farm*, 1983).

In traditional[2] or authority-centric structures, power does not hide; it directs, it commands. It is the non-high-status actor who must obscure: not necessarily someone "low" in absolute terms, they are anyone placed in a relation where they do not hold the higher ground. Their concealment is calculated, protective, and socially expected though harshly sanctioned when discovered.

This concealment should not be mistaken for passivity. It is a proper mode of agency and a tactical withholding of motive. This is a reluctance to expose one's reasoning to a framework that would weaponize it. As Wagner (2012) shows in his account of an undocumented African migrant detained in Israel, his refusal to speak was not a pathology but a strategic act of procedural evasion. The man could not be deported because he could not be identified.
His silence imposed a standoff that forced a foreign State to tolerate his presence without being able to ascertain its legality. But had the man's country of origin been confirmed, had the opacity of his title been lifted, the consequences for him would have been punitive. His concealment was both his shield and his trap: tolerated as long as it worked, punished the moment it failed.

This logic defines many asymmetrical interactions: those in subordinate positions must obscure their motives not because they lack some, but because they know better than to reveal them. They grasp intuitively that truth, when exposed without protective guarantees, is detrimental and triggers punishment. Their opacity is neither weakness nor confusion; this is a deliberate act of cunning.

In more liberal and individualistic configurations, the reverse emerges.
It is the actor in a position of formal authority or hierarchical advantage who attempts to blur their logic in order to escape exposure, avoid justiciability, and preserve their ability to affect others without confrontation. Institutional authority in progressive environments often creates opacity by design.

As of the 1910s, the foundational "iron law of oligarchy" (Michels, 1911/2001) observed that even democratic organizations tend toward the consolidation of power, with leaders hedging against challenge by controlling procedural access or informational flow. Pasquale (2015) extends this, showing how financial companies deliberately obscure their operational logic behind algorithm complexities to shield decision-making and thus reduce constatability. In these cases, transparency is a strategic tool deployed disingenuously.
Furthermore, the transparency discourse itself can be a governance technique whereby sincere actors perform openness and unwittingly strengthen knowledge asymmetry within the system (Birchall et al., 2022).

---

[2] I use 'traditional' as a deictic term, not as a chronological reference to the past. It designates a retrospective stance adopted in the present toward 'what was in the past,' regardless of how distant or how well substantiated that past may be.

The goal is immunity, since legal and procedural frameworks in these settings include legal mechanisms to sanction overt arbitrariness i.e., outcomes that are too blatantly traceable to an illegitimate, self-serving cause. To enact power with vocal or open clarity is to risk being labeled tyrannical and thereby becoming justiciable. As a result, people adapt: the more power they hold under liberal structures, the more likely they are to proactively conceal any trace of negative effect their power may have on others, whether by abuse or even by the fair exercise of their statutory upper-hand.

## 2. Arbitrariness in Social Interactions: Affecting Without Normative Legibility

### 2.1. From Lack of Motivation to Polemological Blockade

#### 2.1.1. Normative Illegibility as Strategic Opacity

Building on insights from information systems and legal theory, we may describe social acts—interpersonal, organizational—as producing reliable consequences—status change, marginalization, reward, harm—without confronting the normativity that would justify them.

The absence of motivation generates a 'why-opacity', a structural ambiguity about causality and intent. The substance of a socially arbitrary act resides in its unjustifiable deviation from the expected norms that govern a given interaction. This deviation may manifest as an infra-normative infraction, transgressing unwritten standards of civil relationship such as loyalty or good faith. Alternatively, it can be a supra-normative infraction, violating more clearly stated expectations like a "given word" confirmed by witnesses, specific contractual clauses, or even a formally established legal status. A deviation may also be transversal, encompassing both infra-normative and supra-normative elements.

The observable form of this arbitrariness is characterized by vague statements, unmotivated decisions, and acts whose inherent logic is unjustified, representing, in essence, the deliberate opacity of the causal structure underlying the behavior.
It is imperative to understand that this opacity does not imply an absence of logic. On the contrary, opacity suggests that the true logic has been intentionally obscured because its exposure would reveal a logic too removed from the normative fabric to remain unnoticed or clandestine. This intentional obfuscation serves strategic purposes, aiming to avoid direct confrontation, evade accountability, and preclude contestation.

This is not a matter of unclear speech or informal behavior. It is what might be called the structural sealing of a scene: the insulation of an act from the conditions of its own rational defeat. Bauman's account (1977) of ritual performance illustrates how linguistic stylization and social dramaturgy function as relational camouflage, redirecting attention from intent to decorum, thereby disarming resistance in advance.

Consequently, the lack of observable logic or constatability for a given social deed renders it difficult to contest. Then, the lack of contestability is profoundly significant, as it allows the social deed and, crucially, its social effects to persist and remain unchallenged as instigated by "Conflict Lateralization".

### 2.1.2. Conflict Lateralization and The Blur of the Wolf Drowned in the Fish

Conflict Lateralization describes the maneuver by which a social actor diverts the potential for frontal conflict, channeling it into a non-justiciable domain to avoid being defeated or corrected. This strategy often leverages tactics such as pseudo courteous language, ambiguous formulations, undeclared but effective decisions, implicit or cavalier consents, and the use of institutional rituals as relational masks.
It resembles a blurred semiotic blend, what I have elsewhere called *le flou du loup noyé dans le poisson* (the blur of the wolf drowned in the fish): a foggy fusion of non-contradiction and camouflage that neutralizes accountability through ambiguity. The ambiguity lies not in the richness of meaning; instead, it is the resistance of even the emergence of a single plausible reading of the behavior, no coherent interpretation maps onto the interaction.

This expression emerged from the analysis of a professional exchange in which apparent collegiality masked an act of discrimination (see Appendix A).
After a researcher openly invited three doctorate candidates to propose contributions to a science-fiction congress, another researcher, in her capacity of PhD supervisor for Fred[3], doctorate candidate #3, talked him out of participating in the event by suggesting it was reserved for the others two PhD students. The reason given? A fog.
This was not an excess of precaution or an innocent clarification. It was a targeted, preemptive exclusion, conveyed through blurry and indirect language. The wolf, representing the adversarial motive to block Fred, was shrouded in a blur and drowned in the fish. The fog veils intent, the fish carries it harmlessly, and the drowning seals the scene. In French, *vague* means both "vagueness" and "wave", underscoring the moving and liquid format of what is vague.

As an intended act of discrimination, it effectively barred Fred from joining and enjoying the playful event organized by his university. It happened for no sensible reason and with no liability, therefore, this act was absolutely arbitrary as herein defined. Through a linguistic breakdown, Appendix A reconstructs the full matter, showing how immotivization, a mechanism whereby someone deliberately, conscious or no, obfuscates their motives within an interaction, produces adversarial effects through inconstatable logical positioning that is immune to contradiction and that lateralizes the conflict.

> **Appendix A: The Blur of the Wolf Drowned in the Fish –** *The Seminal Case Study of Conflict Lateralization*

---

[3] "Fred" is an alias.

When a social actor withholds the rationale for their decision, or articulates it through vague, ambiguous, or ambivalent forms, they do not neutralize the conflict. They deflect it. The conflict is neither defused, nor resolved. On the contrary, it accumulates potency by remaining latent while continuing to generate adversarial, passive-aggressive, and harmful effects.
This is symptomatic of Conflict Lateralization, this strategic avoidance of rational confrontation through which the logic of an act is rendered inconstatable, and consequently incontestable. Its interpersonal outcomes borrow from that immunity, but with a slight shift in presentation: though equally shielded from any recourse or remedy, the effects are real, impactful, and constatable whereas their source is not. The asymmetry of the equation is preserved.

### 2.1.3. Motivation as Polemological Interface and Entropy Reduction

The true function of *motivation* is not primarily to justify the content of a decision.
Rather, its foundational role is to render the decision's underlying logic constatable, that is, observable, interpretable, and available for intersubjective contestation. It is this constatability that opens up the act to rational critique, that makes it vulnerable to logical or normative defeat. Motivation, then, is not the anchor of legitimacy: it is the *opening* to dispute.
It transforms a social effect from an enactment into a proposition, something that can be contradicted. Its purpose is not to justify, but to render the act *attackable*.

At this point, the parallel with the legal system solidifies as a standalone structure.

The same triadic sequence holds:
**Motivation → Constatability → Contestability**

This methodical progression from motivation to contestation, is what conditions condition access to justice. The absence of motivation or the impossibility to reach the constatation phase is rarely accidental. It reflects a defensive strategy deployed to preempt the triggering of a regulated opposition. This is not a polemical game of playing at war; it is a polemological stake, a war at play.

This logic might appear procedural while, in fact, this logic is informational. *Surprisingly*, it can be formalized using Claude Shannon's mathematical theory of communication: a scene becomes justiciable only where epistemic entropy, the uncertainty about the adversary's motives and logic, can be systematically reduced.

Shannon's classic entropy equation

$$H(X) = -\sum p(x) \log p(x)$$

describes the expected uncertainty associated with a message X, based on the probability distribution $p(x)$ over its possible states.

The polemological field functions analogously: justice requires that the motivations underlying an act or position be sufficiently exposed to make their logics ascertainable (constatable) and therefore challengeable (contestable).

Without this informational openness, contest becomes impossible; the scene degenerates into opacity: outcomes emerge without exposable causes, and effects remain insulated from counter-argument.
Motivation, constatability, and contestability together define not only a legal architecture but an epistemic one, a structure that permits justice precisely because it enables uncertainty reduction. The ultimate condition for justice does not lie within law alone, but embedded in the domain of polemology.

The polemological field can be defined by its goal: the deliberate defeat of a prior and adversarial state of fact through an identifiable battleground, a binding opposition.
It can also be delimited by two strict criteria: (1) the declared intention to defeat the adversary (i.e., all the adversaries must know that they are targeted and by which opponent), and (2) the exposure of one's logic to confrontational risk (i.e., the opposition must incur, for all the parties, the possibility to be defeated by the move or response of their opponent); any process failing to meet these conditions, regardless of its intensity or outcome, falls outside the polemological domain.

The term is not reducible to the study of violent geopolitical conflict (Bouthoul, 1951).
Here, it names a synallagmatic relational field: a space in which two or more parties engage in a relationship of intrinsic opposability, governed by minimal rules of engagement, and structured by a fundamental demand for reciprocity in risk exposition.
In contrast with Habermas' ideal of rational consensus through communicative symmetry (Habermas, 1992), this model defines justice as a function of structured contestability, not agreement. The goal is not understanding, but the defeat of a prior state of fact under conditions of mutual exposure. This is exemplified by the principle of adversarial debate, which instead of assuming agreement, assumes non-concealment: the refusal to insulate one's logic from critique. Unlike polemic, which often dissimulates logic behind affect and persuasion, polemology requires its exposure.

Only the polemological field makes a logic exposable and therefore contestable. It locates the arena in which a counter-logic, a counter-act, a counter-argument, can provoke the symbolic or material reconfiguration of the initial state… by concretely, pragmatically defeating it.
As Kojève (1947) observed, the confrontation necessary for recognition must entail risk; in polemological terms, the possibility of defeat is what renders a logic publicly visible and structurally justiciable.

This need not be a formal setting: an intellectually rigorous debate, a professional disagreement, or a vivid keyboard feud among peers all qualify as polemological engagements, provided the logic is not shielded from confrontation. Nor does it have to be a linguistic contest.

Polemology extends to many historically codified forms of conflict.

Aristocratic duels, though violent, were governed by ritualized rules that rendered the logic of confrontation both legible and finite. Where Girard (1972) reads ritual confrontation as the sublimation of violence, the present model treats it as a structure of justiciability: logic made contestable under the rules of engagement. Traditional warfare, insofar as it differs from asymmetrical or unilateral violence, can likewise be understood as a large-scale polemological engagement, wherein opposing strategies and tactical logics are tested within mutually intelligible conventions and objectives.

The same principle governs public debates moderated by deliberative protocols; legal trials structured as codified confrontations in a shared juridical language; and scientific controversies regulated by methodological rules that allow hypotheses to be challenged, falsified, or displaced. Each of these domains expresses the same structural principle: a relation of opposability that permits exposure, contestation, and defeat, and thereby reopens the space for justice.

Where the polemological field is denied or circumvented via Conflict Lateralization, what remains is the effect without the cause.

We can even formalize the modern structure of arbitrariness itself, as a conditional entropy:

$$A = H(L|M)$$

*A* measures the arbitrariness of an act or scene, *L* represents its underlying logic, and *M* the motivations exposed or declared. In a structurally justiciable context, exposed motivation reduces uncertainty about logic; entropy drops. But when arbitrariness prevails, uncertainty about logic remains high despite visible motivations: the act appears explainable yet remains epistemically opaque.

### 2.2. From Sign to Sense: The New Arbitrary Scene

#### 2.2.1. The Interpersonal Settings of Meant Effects

Arbitrariness, in the Saussurean sense, originally concerned the internal structure of the sign, a stable relationship between signifier and signified not grounded in derivational logic (Saussure, 1916/2011). What we are now witnessing is a shift in the unit of analysis from the sign to the sense: the new scene of arbitrariness (Keane, 2003; Silverstein, 2003; Latour, 2005).

While the *scene* provides the material stage, sense is the 'where' from which the logic or motivation must be elicited. This paper's central shift is from *sign* to *sense*, not from *sign* to *scene although* the scene plays a pivotal mediating role (Goodman, 1978; Bauman & Briggs, 1990). The scene is not the destination of analysis, but the stage through which sense is enacted and made consequential (Butler, 1997). The scene hosts the enactment; it is **sense** that governs recognition, consequence, and contestability (Lemke, 1995).

In this context, the scene is not a mere event. It is a performed act with subjective and real-life impact, a micro-structure of interaction that modifies status, position, or relation within a given ecosystem (Bauman & Briggs, 1990; Latour, 2005). And just as a sign may function while remaining arbitrary, so too can a scene produce real and recognizable effects while escaping the space of shared interpretation and confrontation (Shannon, 1949; Luhmann, 1984/2004; Lemke, 1995).

### 2.2.2. The Forensic Causes of Arbitrary Human Play

What renders a scene arbitrary is not its injustice, but its indisputability (Gluckman, 1955; Gallie, 1956). It is not necessarily unfair, it is unreachable. It is not sanctionable in the normative sense of deserving praise or blame (Das, 2007); it escapes intersubjective visibility before it escapes any moral judgment (Butler, 2005). A scene becomes arbitrary when its underlying logic is not secured by shared sense-making, but instead is left gaping to the undecidable: equivalently probable causes among hypothetical possibilities.

The scene is pathologically polysemous, not because meaning is missing, but because a plurality of eligible meanings remains unresolved while still achieving the intended interpersonal outcome (Taussig, 1999; Das, 2007; Devitt, 2021). This undecidability stalls interpretation in a state of hesitation, where multiple logics seem equally plausible but none can be confirmed. It generates a form of interpretive paralysis akin to what cognitive-forensic analysts call "causal overload," (Vrij et al., 2010) where judgment is delayed not by lack of data, but by the absence of disambiguating affordances (Kahneman & Tversky, 1983).
This blocks polemological engagement by suspending the ability to frame a contradictable claim (Pennebaker & Graybeal, 2001).

It is not that the act has no logic; it is that any number of logics might fit, none of which are declared, and all of which remain uncertain and hardly ascertainable (Boltanski & Thévenot, 2006). This uncertainty does not solely reflect subjective doubt, it also denotes the structural impossibility of extracting an objectifiable rationale that could be assessed by a third party under conditions of procedural scrutiny.

In legal-forensic contexts, this absence of verifiable motive breaks the chain of justiciability (Simon, 2004). It disables both investigation and adjudication by denying the minimal threshold of traceable intent (Anderson et al., 2005). It has exited the zone of mutual readability in order to evade the zone of regulated meaning, thus contradiction (Lyotard, 1988; Butler, 1997). The scene does not offer itself as a proposition; it does not announce its conditions (Ronell, 2005). It does not subject itself to intersubjective common grounds, the socio-cognitive milieu where an affirmation becomes possible, and therefore where contradiction becomes thinkable: it affects while remaining invulnerable to counter-affect (Foucault, 2003; Gell, 1998).

This brings us back to the Saussurean insight: arbitrary structures are not dysfunctional. They perform with stability. The sign circulates. The sentence works. The command is obeyed. The arbitrary sense alters the relation. But when the internal logic is unmotivated, when no rationale is exposed or available for confrontation, arbitrariness becomes interpersonal.

Arbitrariness, here, is no longer linguistic but fully social: a property of enacted scenes rather than of symbolic signs, and a logic of narrative sense-making rather than of signifier-signified correspondence.
What has changed, then, is not the principle of arbitrariness, but its field. We are no longer in the bounded realm of linguistic signs. We are now confronting arbitrary scenes: configurations of interactive and interpersonal consequence that structure behavior while evading intelligibility.

### 2.2.3. The Arbitrary, the Social and the Justiciable

In its original phrasing[4], my argument proceeds as follows:
*La motivation est l'interface entre, d'une part, un acte affectant l'interaction au sein d'un écosystème et, d'autre part, l'espace intersubjectif dans lequel sa logique pourra être combattue, répliquée, critiquée, contredite ou révoquée.[5]*

Motivation is not a justification in the moral sense. It is the technical module through which a deed becomes intersubjectively accessible and polemologically vulnerable. Without it, the sense of the scene cannot be accessed, only endured.

This is precisely where Conflict Lateralization begins to operate. *L'acte latéralisé empêche d'en voir le chemin. C'est ainsi que l'arbitraire ne naît pas du vide, mais du langage immotivisé produisant néanmoins des effets concrets.[6]*

An arbitrary scene is not chaotic. It is highly structured. But its structure is sealed; it is a form of causality that affects without being traceable, a sentence that commands without being semantically opposable (Cotterill, 2003).

---

[4] Such original phrasing is intended to share with the reader some of the princeps materials grounding this framework, originally developed in a non-English language. Their inclusion preserves the conceptual nuance and idiomatic structure of the argument as first articulated, before its translation into English for this publication. This gesture acknowledges that theoretical inquiry does not emerge exclusively in English and that certain formulations retain a distinctive clarity in their language of conception.
[5] Motivation is the interface between, on the one hand, an act that affects interaction within an ecosystem, and on the other hand, the intersubjective space in which its logic can be challenged, countered, critiqued, contradicted, or revoked.
[6] The lateralized act prevents one from seeing its path. This is how arbitrariness does not arise from a vacuum, but from immotivized language that nonetheless produces concrete effects.

In paralegal terms, this is akin to tampering with evidence by preventing the act from being investigable at all (MacCormick, 2005). The logic is not so much made invisible than it is intentionally preempted, as if the scene were tucked in, the doorway shut, with just enough motivational red-tape to forestall any reactive response or reconstructive entry (Twining, 1999). In cognitive terms, this creates what might be called a hermetic context, where the agent pre-engineers the impossibility of response by cloaking the very conditions of contestability (Grice, 1975; Schauer, 1991).

Arbitrariness, here, is not failure: it is design.

This reshuffles the criteria of authority. It is no longer the power to act without rules or to enforce inequitable decisions. It is the power to act without justiciability. Arbitrariness is the capacity to bind without exposing one's logic to any potential contradiction.
Hence the polemological essence of justice. *Ce champ polémologique commence là où la logique de l'acte est constatable. La possibilité de défaite (du fait antérieur) devient alors la vraie unité de mesure d'une scène juste.*[7]

Justiciability is not a procedural ornament. It is the mandatory precondition for justice.
And justiciability is not a legal construct, but a polemological one. An interaction is not *just* because it is kind, coherent, or even lawful. It is *just* because it produces outcomes that proceed from logics that can be confronted, argued, and defeated.
Where that possibility is blocked, where motivation is refused and authorship is evaded, what remains is still an act, still a consequence, still a fact. But it is no longer legible, no longer liable. It is also no longer challengeable; and no longer accountable. It is, however, in the strongest sense of the term: arbitrary.

***

**Conclusion**

This paper has proposed that arbitrariness is not an aberration, a failure, or a deplorable flaw of our collective symbolic structures. It is a socio-cognitive embedded mechanism: one that enables operability without justification, consequence without contradiction, and impact without exposure. By shifting the analytic unit from the sign to the sense, and by situating the scene as a site of enacted yet unmotivated consequence, we have outlined the intemporal grammar of arbitrariness that branches far beyond language in itself and permeates to our legal, political, and everyday interaction.

In this framework, opacity is not the sheer absence of clarity. It is a technique of insulation, a protective boundary that seals a logic away from confrontation.

---

[7] The polemological field begins where the logic of the act is ascertainable. The possibility of defeat (of the prior act) then becomes the true unit of measure for a just scene.

What makes this insulation powerful is precisely its capacity to affect without being answerable, to shape decisions and outcomes while preempting their disputability.

Transparency, in disclosing motivation and triggering constatability, is not a virtue. It is an epistemic prerequisite for justice to even be thinkable and for coordination to be actionable regardless of trust between the interactive parties.

When the logic of an act can be declared, situated, and contradicted, the scene unlocks its polemological potential and, therefore, becomes socio-politically accountable. The problem is not the existence of arbitrary structures, but their ability to circumvent justiciability while prompting detrimental effects for the individual at the receiving end.

This is why we ought to think beyond normative expectations of fairness or goodwill.

What this theory calls for is a structural ethic of contestability. A redesign of communicative, legal, and institutional interaction around the obligation not only to act, but to render one's action epistemically exposable. In many settings or critical turns, the constatability of motive in interpersonal action must be recognized as a precondition for good-faith legitimacy.
In this regard, many micro-relational configurations that generate significant real consequences without disclosing their underlying logic must be recognized for what they are: scenes of motivated unaccountability within a suspicious arbitrary background.

As I have argued throughout, arbitrariness is not failure, it is design. It is the grammar of an authority that adjudicates alone but for you, and whose sentence knows no appeal. And as such, it must be confronted not with moral stance, but with chronological accounting: the one that recounts what happened in order to hold the author accountable… whenever an actual prejudice has occurred.

I admit that this is not easy. Forensic reconstitutions are indeed warranted.

It is not the absence of motive that renders a deed arbitrary. It is the refusal to make it constatable. No witness? No crime. No articulated account of the loss? No investigation or redress proceedings can even be initiated.

In language, arbitrariness is the normal condition of signs. In justice, it is what prevents an act or a situation from entering the perimeter of justiciability. In interpersonal relationships, it is what allows an effect to remain unchallenged, uncorrected, and unrepaired, not because it is necessarily unjust, but because it is structurally unaccounted for.

However, is this always nefarious? Is it accurate or even analytically useful to presume that all arbitrary acts are precursors to harm, silence, or social abuse? Not even close.

Until now, the mechanism whereby certain acts become inaccessible to replication, response, or rational opposition has remained emotionally over-labeled and systematically under-theorized. It lacked an intelligible framework that could connect linguistic opacity with interpersonal consequence.
I hope I have managed to fill this conceptual gap by showing that arbitrariness describes a logical-relational asymmetry, not a personality or institutional flaw.

If we are to take arbitrariness seriously as a structural condition, then we must recognize it in all its valences around the whole spectrum of its interpersonal outcomes, for, if arbitrariness is a semiotic tool of social functionality paired with an engine of domination, it is nonetheless the carrier of non-instrumental care.
Sometimes, what resists explanation is not meant to deceive.
Sometimes, it is not meant to be claimed.

In this model, arbitrariness is not the opposite of justice or fairness but the failure to share what makes it possible—motivational legibility.
When social critique conflates arbitrariness with domination, it loses sight of a distinct kind of harm: one that is neither coercive nor ideologically imposed, although still rooted in epistemic exclusion. This is harm not through oppression but through opacity, by denying a person access to the rationale behind consequential acts or by restricting access to critical information.
It deprives them of their capacity to meaningfully respond, react and contest.

The lawsuit filed in March 2022 by Gabby Petito's parents against Brian Laundrie's offers a textbook case of arbitrariness as theorized here.

After learning of Petito's daughter's passing (her remains were discovered on September 19, 2021), the Laundries officially voiced their hope while refusing private communication, as a stance loaded with clandestine hostility toward a distressed and ill-informed interlocutor. Their opaque behavior had profound effects: it disarmed the Petitos, delayed their response, and deepened their state of anguish.

The Laundries' demeanor was a deliberate act of immotivization: outward behavior that induce clear interpersonal and legal consequences, coupled with a commitment to not disclose the rationale behind it. Moreover, the Laundries engaged in Conflict Lateralization, making direct confrontation impossible by channeling information through vague public statements.
The result was epistemic exclusion: the Petitos were structurally prevented from contesting, reasoning, or even meaningfully interpreting the Laundries' conduct because its motivation was never made constatable.

In this way, the case illustrates arbitrariness not as domination, coercion or ideology, but as a structural refusal of motivational legibility, a creepy and awkward performance meant to keep the parties just below the bar of justiciability for the relevant period of time before it inevitably devolved into an open conflict of its own right.

In February 2024, the families reached a confidential settlement suggesting that for once, the harm effectively inflicted throughout tactical obfuscation, the time and chances effectively wasted by playing on asymmetrical warfare were sufficient to trigger a legal remedy.

However, opacity is not always harmful.

The same structural condition that conceals motivation and disables contestation can, in other contexts, enable profound acts of care and protection.

This is precisely what social critique tends to overlook when it treats all opacity as domination. It forfeits the ability to detect a distinct kind of good: the capacity to affect another person positively without explanation, through acts that are benevolent, protective, or loving without offering their own justification.

This is not only what is expected in foundational social bonds (such as parenting, where "love" must take form in unspoken action), but also what quietly structures many of our everyday social interactions: the strict teacher who, invisibly, shields a student from harm; the colleague who protects without claiming credit; or the Severus Snape who saves without disclosure. These acts, too, are arbitrary in structure. Yet, they cannot be reduced to domination, ideology, or violence. They are non-contestable goods, just as others are non-contestable harms.

To reclaim arbitrariness as a neutral operator is to recognize that humans are not only capable of arbitrary cruelty, but also of arbitrary grace. The rationale behind such benevolent acts, often attributed to "love," "character," or "generosity", hardly unfold any logical explanation. "Love" may name the affect, but it does not justify the gesture. It does not make it reproducible, transferable, or answerable. Why would anyone nurture and cater to another non-relative person, inconspicuously, without asking for anything? No one knows. It is not performative, and it is not claimed. It just happens.

This opacity is not a flaw. It is the structure. As Shannon showed, a message can transmit without semantic congruence. As Barthes taught, a myth can structure belief while masking its construction. So too can an act transform another's life, without revealing the logic behind it. Arbitrariness, then, can no longer be described as a latent tool of exclusion without also being introduced as the hidden conduit of unclaimed protection. It carries out harm but it also carries out care, both by structural inaccessibility.

Is arbitrariness the unassailable cloak of pervasive abuse? Not really. Most times, it is the quiet language of our humanity at its best: it simply gives without declaring itself, it also nurtures without intent to be noticed.

***Arbitrariness Beyond Domination & The Human Grace Beyond Invisible Motives***

# **Appendix A**: The Blur of the Wolf Drowned in the Fish
## *The Seminal Case Study of Conflict Lateralization*

The following Case Study is based on real-life materials. For privacy concerns, all identifying details have been anonymized and altered. The analysis focuses on the structural properties of the interaction. The original French conversation is provided in the footnotes.

—

A.1 Context and Discursive Act: The Interpersonal Scene

In spring 2025, a call for contributions to the Science Fiction Congress (SFC), a one-week event organized by a French university for all its staff members (scientists, students and administrative agents alike), was shared by Researcher A and addressed collegially to three PhD candidates and Researcher B, all five working within the same lab research.

The first message of the thread was inclusive.

Step 1: The Initial Invitation (Researcher A to All)

> Hello everyone,
> Great, thank you for your availability [for Monday's PhD meeting].
> Attached is the SDF a call for papers; could you please review it and give us some suggestions/proposals that we can discuss on Monday?
> Best regards.[1]

Step 2: The Preemptive Reframing (Researcher B to Fred)

Immediately after this invitation, Researcher B, the supervisor of Fred, PhD Student #3, replied privately to the latter, preemptively framing the SFC contribution as both nonessential and peripheral:

> I will be happy to listen to you on Monday regarding the information you've gathered for the country comparison (if you encountered any difficulties, how you overcame them...).
> It will also be a question of hypothetically participating in the SFC.
> It's a bit far from the Bacchus project that your PhD is focused on, but perhaps the PhD projects #1 and #2 would like to propose something.[2]

---

[1] *"Bonjour à tous,*
*Super, je vous remercie de votre disponibilité [pour la réunion PhD de lundi]. Voici, en PJ, un appel, pouvez-vous le consulter et nous faire des suggestions/propositions que nous discuterons lundi ?*
*Bien à vous."*

[2] *Je t'écouterai avec plaisir lundi, sur les infos glanées pour la comparaison entre les pays (si tu as rencontré des difficultés, comment tu les as surmontées...).*
*Il sera aussi question de participer éventuellement à la SFC.*

In her message to Fred only, Researcher B repurposes the previously collective opportunity as conditional, secondary, and reserved for the other two PhD candidates.

The PhD project relevant to Fred is positioned as irrelevant to the SFC, in spite of the Congress being established precisely to offer relief from official research programs, by promoting imaginative exploration through a playful symposium that welcomes speculative contributions from all scientific backgrounds. The implicit exclusion is not stated but softly suggested by attenuation ("hypothetically", "a bit far from the Bacchus project", "perhaps").

Already, the field is undermined through tonal diminishment.

Fred responds openly, with both enthusiasm and a heart-wrenching premotion of the rejection his participation in the Congress will trigger:

> Regarding the SFC, I'm quite excited because I really like science fiction. But you're right, if I submit a conference proposal, the topic will have no connection to Bacchus, other than AI, so you might not approve. In any case, I have a thousand ideas in mind ^^.[3]

Step 3: The Final Reply or The Last Straw (Researcher B to Fred)

Caught off guard by Fred's authenticity, Researcher B replies again, visibly uncomfortable, with a highly constructed yet evasive formulation:

> For the SFC, you might give some ideas, but the valorization of lab's projects must be prioritized. If the other projects don't respond, then there might be grounds to see if 'off-project' ideas can be considered.[4]

→ As an external observer of this scene, it was this final reply that crystallized the tension that had been at play since the beginning. The lack of descriptor urged me to coin the expression that immediately came up to my mind: *"le flou du loup noyé dans le poisson"* i.e., "the blur of the wolf drowned in the fish" (the "Wolf's Blur" or the "Wolf's Fog").

A.2 Pragmatic Structure of the Final Reply: The Linguistic Immotivization

Each clause of this response is syntactically correct, yet pragmatically evasive. Its function is not to resolve, clarify, or assign. It is to seal the space of initiative while avoiding exposure to direct contradiction.

---

*C'est un peu loin du projet Bacchus qui intéresse ton PhD, mais les projets PhD #1 et #2 voudront peut-être proposer quelque chose.*

[3] *Pour le SFC, je suis assez emballé car j'aime bien la science fiction. Mais effectivement, si je soumets un projet de conférence, le sujet n'aura aucun rapport avec Bacchus, si ce n'est l'IA, donc tu ne seras peut-être pas d'accord. En tous cas, j'ai mille idées en tête ^^.*

[4] *Pour le SFC, tu pourras donner des idées mais il faut prioriser la valorisation des projets pour le labo. Si les autres projets ne répondent pas, il y aura matière à voir si des idées 'hors projet' sont envisageables.*

A.2.1. Analysis of the Final Reply's Pragmatic Structure

| CLAUSE | FUNCTION | MECHANISMS | EFFECTS |
|---|---|---|---|
| "you might give some ideas" | Priming a no-go clearance, granting an open no(n)-permission | Uncertain modal of future action ("you might") instead of a modal anchored in the indicative present (e.g., "you can").<br>The recipient of the ideas is undefined, leaving the question: "To whom might Fred's ideas be given?" | Simulates a possibility while incapacitating Fred's ability to act. |
| "but the valorization of lab's projects must be prioritized" | Binding the no(n)-permission to another unspecified priority | Uses the conjunction "but" to mark the opposition between the no(n)-permission and the priority to be attended to first. Employs the impersonal French construction *"il faut"* (a strong, agentless obligation), which is translated into English using the modal "must" in the passive voice. This choice reflects the softening of the injunction by removing any human subject from the sentence: "the valorization of lab's projects must be prioritized."<br>To be clear, neither Researcher B nor his PhD Student #3 knew what tasks "the valorization of lab's projects" was specifically referring to, as all three PhD projects under discussion were conducted within the same lab. | Subordinates the fulfillment of what is not permitted to a set of unspecified lab-related tasks without sounding like a coercion. |
| "If the other projects don't respond..." | Adding another layer of conditionality to a no(n)-permission to be acted once unspecified priorities will be complete | Introduces a third-party contingency without naming the people involved. Instead of saying "If PhD Student #1 and #2 don't reply", Researcher B designates these individuals by their projects' name. Besides being dismissive, this metonymy aims to obfuscate the differential treatment between the two PhD candidates who can reply to the call for papers and Fred who cannot. | Displaces agency onto the silence of the unnamed PhD Students #1 and #2 who, them, for some unknown reason, have the right to participate to the SFC. |
| "then, there might be | Submitting Fred's participation to | Deploys a tautological placeholder. "There might be grounds to see" is used instead of a direct commitment | Restricts any forward movement by Fred who |
| grounds to see..." | another checkpoint in the event the other exclusionary clauses preventing him from joining the SFC are cleared. | like "We will see". The phrase "There might be" is motionless with no human subject and uses a modal of uncertainty to replace a clear projection of what will happen "then". The verb "…to see" is purely contemplative, ethereal and on hold, with no intent to either make a decision or cause any form of action. | cannot be sure of what is authorized for him even after the conditions of his "no(n)-permission" are technically lifted. |
| "...if 'off-project' ideas can be considered" | Further delaying and distancing Fred's unlikely response to the Congress call. | Creates a double delay with the "if" conditionality and the unpersonal and passive voice of "can be considered". "Off-project ideas" is used to replace "Fred's ideas", another metonymic attempt to avoid pointing at him too obviously (although the negative "off-project" qualifier gives away, by association, how Fred's is looked upon by his supervisor). | Renders the very possibility of Fred's participation contingent on a hypothetical permission and undefined future consideration. |

A.2.2 From Linguistic Form to Social Force

Researcher B's communication style is the foundational illustration of immotivization: speech that conforms to grammatical rules but refuses to anchor in an exposable logic. It generates effects without committing to causes. The invitation becomes a procedural delay; the enthusiasm becomes structurally sidelined. And yet, this reply achieves its purpose: it reinscribes control, reinjects hierarchy inside of the SFC's horizontal initiative, and expels Fred from a space that Researcher A had officially opened to him and the other PhD Students… without ever issuing an explicit refusal!

This discursive technique does not argue. It absorbs. It abusively reclaims authority by diffusing it in a procedural haze. The dialogue reveals a sophisticated form of procedural warfare, a kind of static moving of the goalposts. The supervisor's tactic is impressively remote and anticipatory; she constructs an entire maze of obstacles before any step is even in motion, using the pure force of language to foreclose any possibility of action. Each new condition is presented as an unchangeable rule, yet the set of rules is constantly in motion, creating a disorienting and ultimately impassable labyrinth. This effect is achieved through what can only be described as layer after layer of (re)movals: the active removal of Fred's opportunity and agency is perfectly camouflaged by the constant (re)moving of the conditions of polemological engagement.

It is this dense mille-feuille that drowns the "wolf" of intent in the "fog" of plausible deniability, making it a perfect execution of Conflict Lateralization.

## A.3 Naming the Fog: Le flou du loup noyé dans le poisson

- Where this terminology comes from

The expression "le flou du loup noyé dans le poisson" (the wolf's blur drowned in the fish) is a conceptual portmanteau born out of this Case Study, from a spontaneous fusion of two potent French idioms into a single, unique and specific one. It inherits its structure from two distinct historical strata of French parlance.

The first, *"noyer le poisson"* (to drown the fish), is a 19th-century phrase derived from fishing tactics, which describes the active, intentional creation of confusion to evade a direct question. The second, *"quand c'est flou, c'est qu'il y a un loup"* (when it's blurry, there's a wolf), is a more modern piece of good sense popularized in 21st-century political discourse, which voices a deep suspicion that ambiguity is being used to conceal a hidden, malicious problem. The spontaneous blend of these two captures the PhD supervisor's maneuver: it is the creation of a "blur" to mask a malevolent "wolf" inside of fishy waves of semiotics. The phenomenon herein displayed is a very common trick that was missing its own entry, hence the formation of this expression as I was parsing the conversation between Fred and his PhD supervisor. The metaphor objectivizes this experience where the wolf represents the longing to block or control, the fish is the impersonal neutral-looking form under which the predation is disguised; and the blur is the surface effect: vague, shallow, under-specified, indeterminate, yet strategically targeting something that is hindered, sabotaged or undermined.

It is not that the wolf is absent: it is that it has been drowned by what could be called grammatical liquidity. Here, the attack on the meaning is carried out through a linguistic device, however, the scene of an enacted sense is always social when interpersonal: immotivization can be located in silence instead of words, or in unsettling sequels of behavior with or without diluting sentences. This leads to a deed with no declared goal and no official opponent, yet an act that still causes the other individual at the receiving end to fail.

It is pervasive and no relational setting is spared, be it at work, in the street, with friends, family, or random individuals committed to extract something from someone else out of a fog-like apparatus.

- Why this terminology is useful

This Case Study is the occasion to shed light on key features that separate the Wolf's Blur from what is commonly referred to as Word Salad. Although both involve a degree of linguistic opacity, World Salad functions primarily as an artefact of internal confusion from the speaker and is mostly displayed orally instead of in writing. It produces a flow of apparently coherent phrasing, sometimes nice sounding but is actually a nonsensical flow of utterances with no clear directed external aim. The impact for the other person may be one of disorientation or superficial admiration for rhetorical intensity, but it is not geared toward a pragmatic end result on how the interlocutor should react or not react.

In contrast, the Wolf's Blur signature is to engage opacity with precision: it relies on coherence and plausibility to obtain a highly specific effect, response that serves the speaker, in addition to the paralysis of their interlocutor's sense-making engine.

For instance, here, Researcher B's language was in pursuit of one circumscribed goal: Fred's non-participation to the SFC. This form of immotivized language does not reflect a disorderly fluency of the speaker, but the selective suspension of clarity in service of control. As such, this notion constitutes a standalone interpretive category, analytically distinct from disorganized speech patterns and is essential for honing refined interpersonal reading skills.

Another crucial distinction lies in the speaker's alignment with the lack of clear content communicated. Word salad emerges reactively, triggered by the social context that is going to prompt a word salad eruption; though the speech production is deliberate, its meaninglessness is not. By contrast, the Wolf's Blur is, by definition, driven more by a preemptive than a reactive impulse, and it is never accidental: even when the agent lacks self-awareness of their rhetorical strategy, the opacity amounting to the speech deprivation of clear meaning is in line with the individual's wish. Otherwise stated, where the Word Salad producer will feel genuine dissatisfaction and disappointment when realizing that his speech was nonsensical, we can be sure that the Wolf's Blur farmer will feel safe and validated when his statement's opacity (and consequential nonsensicality or, rather, *asensicality*[5]) is confirmed.

To put it in a rare set of psychoanalytic and insightful terms, whereas Word Salad production is ego-dystonic, Wolf's Blur farming is ego-syntonic.

A.4 Arbitrary Implication: The Triadic Sequence

The conversational logs exemplify what is Conflict Lateralization; two parties with no interchangeable roles: one is a soloist underground fighter; one is their punching ball. With Researcher B acting as the fighter, the logic of the conflict is:

- Immotivized: it has been removed, subtilized; hence
  - Inconstatable: it cannot be plainly stated or located; hence
    - Incontestable: it cannot be opposed within any pragmatic frame.

It follows the rule of our triadic sequence. Accordingly, the adversarial effects of the ongoing fight are real and tangible: retraction of opportunity, symbolic marginalization, preservation of hierarchy and… the discriminatory exclusion of Fred.

As an epilogue of this story, because of the unspoken pressure coming from his PhD supervisor, PhD Student #3 will not push back and will simply give up on both submitting ideas and attending to the Congress, whereas PhD Students #1 and #2 will be explicitly encouraged by Researcher B to draft and send their suggestions to the SFC organizers.

Indeed, the scene of this Case Study had no entry point for contradiction; on the surface, zero exposure.

---

[5] The neologism "asensicality" relies on the privative prefix "a-," signaling the absence or withdrawal of sense, rather than its distortion or contradiction. It is not to be conflated with "nonsensicality," which implies the presence of noise, incongruity, or internal inconsistency. Asensicality, on the other hand, names the hollowed-out residue left when sense has been redacted or suspended; a structured vacancy that still operates within a communicational frame, but no longer allows for meaning anchoring, readable motivation, and possible contradiction. It is a silence that speaks in form, not in content.

This was not failure. It was design.

Also, the whole conversation epitomizes Conflict Lateralization, using linguistic immotivization as a magic shield against accountability. In our everyday life and casual interactions, not only such magic shields are rampant, but they are ridiculously effective for they prevent the parties from entering inside of the polemological field thereby blocking them access to justiciability.

This is textbook manipulation with the interface of opacity as authority: arbitrariness.

A5. Arbitrary Bonus: A Visual Sensing

Figure A5.1
*The Full Blur Illumination*

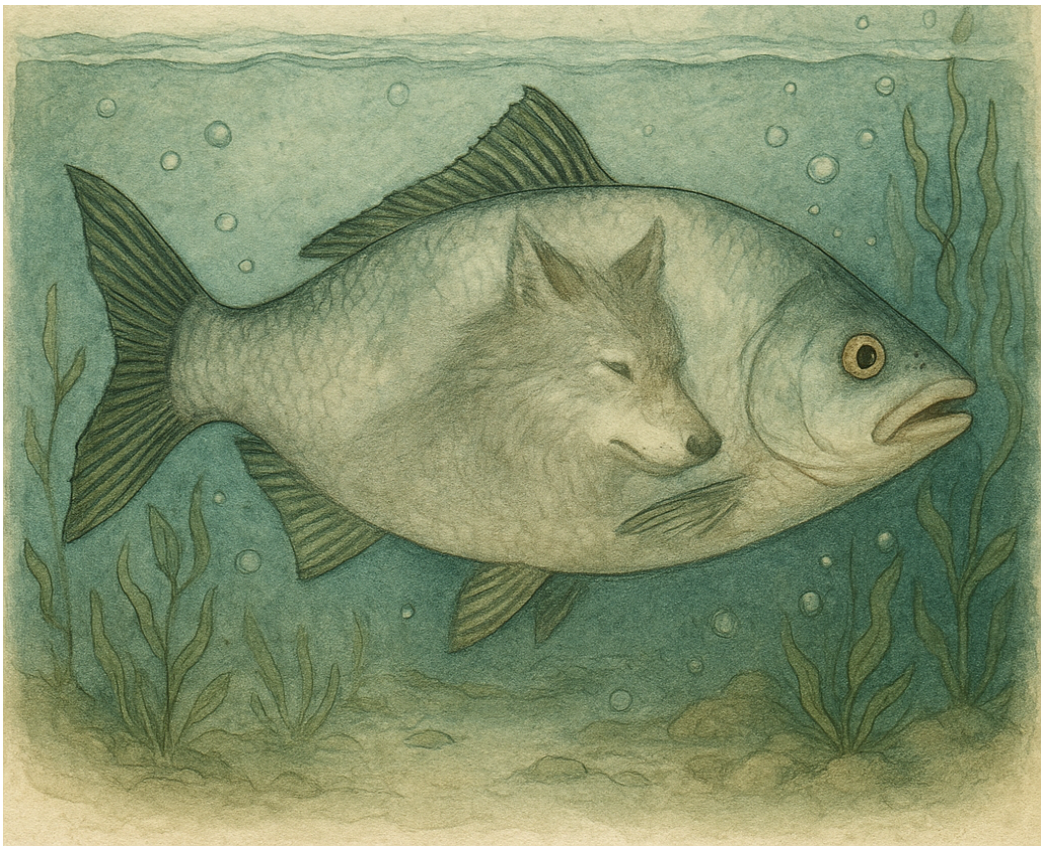

*Note. AI-generated image created using a text-to-image model (OpenAI, 2025), illustrating the 'blur of the wolf drowned in the fish': the wolf represents predatory agentive lurk, the fish its submarine disguise, and, the blur, the semiotic opacity that drowns and neutralizes any self-preserving reaction — such as contestation.*

<p style="text-align:center">\*\*\*</p>